\documentclass[10pt,twocolumn,letterpaper]{article}

\usepackage{iccv}
\usepackage{times}
\usepackage{epsfig}
\usepackage{graphicx}
\usepackage{amsmath}
\usepackage{amssymb}
\usepackage{booktabs}
\usepackage[protrusion=true,
         	expansion=true,
         	tracking=true,
         	final,
         	stretch=25,factor=1250]{microtype}
\usepackage{multirow}
\usepackage{rotating}
\usepackage{enumitem}

\usepackage[breaklinks=true,bookmarks=false]{hyperref}

\iccvfinalcopy 

\def\iccvPaperID{2195} 

\ificcvfinal\fi

\begin{document}

\title{Learning Adaptive Neighborhoods for Graph Neural Networks}

\author{Avishkar Saha$^{1}$, Oscar Mendez$^{1}$, Chris Russell$^{2}$, Richard Bowden$^{1}$\\
$^{1}$Centre for Vision Speech and Signal Processing, University of Surrey, Guildford, UK \\ $^{2}$ Amazon, Tubingen, Germany \\
{\tt\small \{a.saha, o.mendez, r.bowden\}@surrey.ac.uk, \tt\small cmruss@amazon.com}
}

\maketitle
\ificcvfinal\thispagestyle{empty}\fi

\begin{abstract}
Graph convolutional networks (GCNs) enable end-to-end learning on graph structured data. However, many works assume a given graph structure. When the input graph is noisy or unavailable, one approach is to construct or learn a latent graph structure. These methods typically fix the choice of node degree for the entire graph, which is suboptimal. Instead, we propose a novel end-to-end differentiable graph generator which builds graph topologies where each node selects both its neighborhood and its size.  Our module can be readily integrated into existing pipelines involving graph convolution operations, replacing the predetermined or existing adjacency matrix with one that is learned, and optimized, as part of the general objective. As such it is applicable to any GCN. We integrate our module into trajectory prediction, point cloud classification and node classification pipelines resulting in improved accuracy over other structure-learning methods across a wide range of datasets and GCN backbones. We will release the code.
\end{abstract}
\section{Introduction}
The success of Graph Neural Networks (GNNs) \cite{duvenaud2015convolutional, bronstein2017geometric, monti2017geometric}, has led to a surge in graph-based representation learning. GNNs provide an efficient framework to learn from graph-structured data, making them widely applicable where data can be represented as a relation or interaction system. They have been effectively applied in a wide range of tasks \cite{monti2019fake}, \cite{stokes2020deep} including particle physics \cite{choma2018graph} and protein science \cite{gainza2020deciphering}.

In a GNN, each node iteratively updates its state by interacting with its neighbors,  typically through message passing. However, a fundamental limitation of such architectures is the assumption that the underlying graph is provided. 
While node or edge features may be updated during message passing, the graph topology remains fixed, and its choice may be suboptimal for various reasons.
For instance, when classifying nodes on a citation network, an edge connecting nodes of different classes can diminish classification accuracy. These edges 
can degrade performance by propagating irrelevant information across the graph. \color{black} When no graph is explicitly provided, such domain knowledge can be exploited to learn structures optimized for the task at hand \cite{fatemi2021slaps, chen2020iterative, franceschi2019learning, elinas2020variational}. However, in tasks where knowledge of the optimal graph structure is unknown, one common practice is to generate a $k$-nearest neighbor (\textit{k}-NN) graph. \color{black} In such cases, $k$ is a hyperparameter and tuned to find the model with the best performance. For many applications, fixing $k$ is overly restrictive as the optimal choice of $k$ may vary for each node in the graph. While there has been an emergence of approaches which learn the graph structure for use in downstream GNNs \cite{zheng2020robust,kazi2020differentiable,kipf2018neural}, all of them treat the node degree $k$ as a fixed hyperparameter.



 We propose a general differentiable graph-generator (DGG) module for learning graph topology with or without an initial edge structure.
Rather than learning graphs with fixed node degrees $k$, our module  
generates local topologies with an adaptive neighborhood size.
 This module can be placed within any graph convolutional network, and jointly optimized with the rest of the network's parameters, learning topologies which favor the downstream task without  hyperparameter selection or indeed any additional training signal. The primary contributions of this paper are as follows:
\begin{enumerate}[nolistsep]
	\item We propose a novel, differentiable graph-generator (DGG) module which jointly optimizes both the neighborhood size, and the edges that should belong to each neighborhood. Note a key limitation of existing approaches \cite{zheng2020robust, kipf2018neural, kazi2020differentiable, fatemi2021slaps, chen2020iterative, elinas2020variational, wu2022nodeformer} is their inability to learn neighborhood sizes.
	\item Our DGG module is directly integrable into any pipeline involving graph convolutions, where either the given adjacency matrix is noisy, or unavailable and must be determined heuristically. In both cases, our DGG  generates the adjacency matrix as part of the GNN training and can be trained end-to-end to optimize performance on the downstream task. Should a good graph structure be known, the generated adjacency matrix can be learned to remain close to it while optimizing performance.
	\item To demonstrate the power of the approach, we integrate our DGG into a range of SOTA pipelines --- without modification --- across different datasets in trajectory prediction, point cloud classification and node classification and show improvements in model accuracy.
\end{enumerate}

\section{Related work}
\textbf{Graph Representation Learning:}
GNNs \cite{bronstein2017geometric} are a broad class of neural architectures for modelling data which can be represented as a set of nodes and relations (edges).
Most use message-passing to build node representations by aggregating neighborhood information. A common formulation is the Graph Convolution Network (GCNs) which generalizes the convolution operation to graphs \cite{DBLP:conf/iclr/KipfW17,defferrard2016convolutional, wu2018moleculenet,hamilton2017inductive}.
More recently, the Graph Attention Network (GAT)  \cite{velivckovic2018graph} utilizes a self-attention mechanism to aggregate neighborhood information.
However, these works assumed that the underlying graph structure is fixed in advance, with the  graph convolutions learning features that describe pre-existing nodes and edges. In contrast, we simultaneously learn the graph structure while using our generated adjacency matrix in downstream graph convolutions. The generated graph topology of our module is jointly optimized  alongside other network parameters with feedback signals from the downstream task.

\textbf{Graph Structure Learning:}
In many applications, the optimal graph is unknown, and a graph  is constructed before training a GNN. One question to ask is: ``Why isn't a fully-connected graph 
suitable?'' Constructing adjacency matrices weighted by distance or even an attention mechanism \cite{velivckovic2018graph} over a fully-connected graph incorporates many task-irrelevant edges, even if their weights are small. While an attention mechanism can zero these out --- i.e., discover a subgraph within the complete graph --- discovering this subgraph is challenging given the combinatorial complexity of graphs. A common remedy is to sparsify a complete graph by selecting the $k$-nearest neighbors ($k$-NN). Although this can prevent the propagation of irrelevant information between nodes, the topology of the constructed graph may have no relation to the downstream task. Not only can irrelevant edges still exist, but pairs of relevant nodes may remain unconnected and can lead GCNs to learn representations with poor generalization \cite{zheng2020robust}.

\color{black}
To overcome this, recent works constructed bespoke frameworks which learn the graph's adjacency matrix for specific tasks. For instance, in human pose estimation, some methods \cite{shi2019two,liu2020disentangling} treat the elements of the adjacency matrix as a set of learnable weights. However, as each element is treated as a learnable parameter, the learned adjacency matrix is unlinked to the representation space and can only be used in tasks where there is a known correspondence between training and test nodes. This is not the case for many vision and graph tasks. Others have \cite{kipf2018neural, elinas2020variational, lao2022variational} employed variational inference frameworks to sample the entire adjacency matrix. Franceschi et \textit{al.} \cite{franceschi2019learning} jointly learned the graph structure and the parameters of a GCN by approximately solving a bilevel program. NodeFormer \cite{wu2022nodeformer} and IDGL \cite{chen2020iterative} instead learned latent topologies using multi-head attention \cite{vaswani2017attention}. There are two key differences between these methods and ours. First, we simplify optimization by factorizing the adjacency matrix distribution from which we sample the neighborhood for each node, as opposed to sampling the entire matrix. Second, these methods are bespoke frameworks specifically designed for node and graph classification. They leverage knowledge of the task in their loss functions, such as graph smoothness and sparsity \cite{chen2020iterative}. As these methods are tailored to graph-based tasks only, they cannot be dropped into any GCN without modification, limiting their applicability to non-graph tasks like vision. In contrast, our module is both GCN and task-agnostic, and can be integrated into any GCN pipeline and trained using the downstream task loss.

In contrast to the bespoke frameworks above, recent methods \cite{zheng2020robust,luo2021learning,kazi2020differentiable} took a more module-based approach similar to ours. As these approaches learned the graph structure entirely from the downstream task loss, there is less domain knowledge to leverage compared to methods constructed for specific tasks. Consequently, sparsity is often induced through a $k$-NN graph. \color{black} Here, $k$ is a scalar hyperparameter selected to control the learned graph's node degree.

Unlike these works, we generate neighborhoods of varying size by learning a distribution over the edges \emph{and} over the node degree $k$. Each node samples its top-$k$ neighbors \color{black} (where $k$ is now a continuous variable), \color{black} allowing it to individually select its neighborhood and the edges that should belong to it, in a differentiable manner. Additionally, a known `ideal' graph structure can be used as intermediate supervision to further constrain the latent space.

\section{Method}
Here, we provide details of our differentiable graph generation (DGG) module. We begin with notation and the statistical learning framework guiding its design, before describing the module, and how it is combined with  graph convolutional backbone architectures.


\textbf{Notation} We represent a graph of $N$ nodes as $G = (V, E)$: where $V$ is the set of nodes or vertices, and $E$ the edge set. A graph's structure can be described by its adjacency matrix $A$, with $a_{ij} = 1$ if an edge connects nodes $i$ and $j$ and $a_{ij} = 0$ otherwise. This binary adjacency matrix $A$ is directed, and potentially asymmetrical.

\begin{figure}[t]
	\centerline{\includegraphics[page=5,trim=0 580 0 15,clip,width=1.\linewidth]{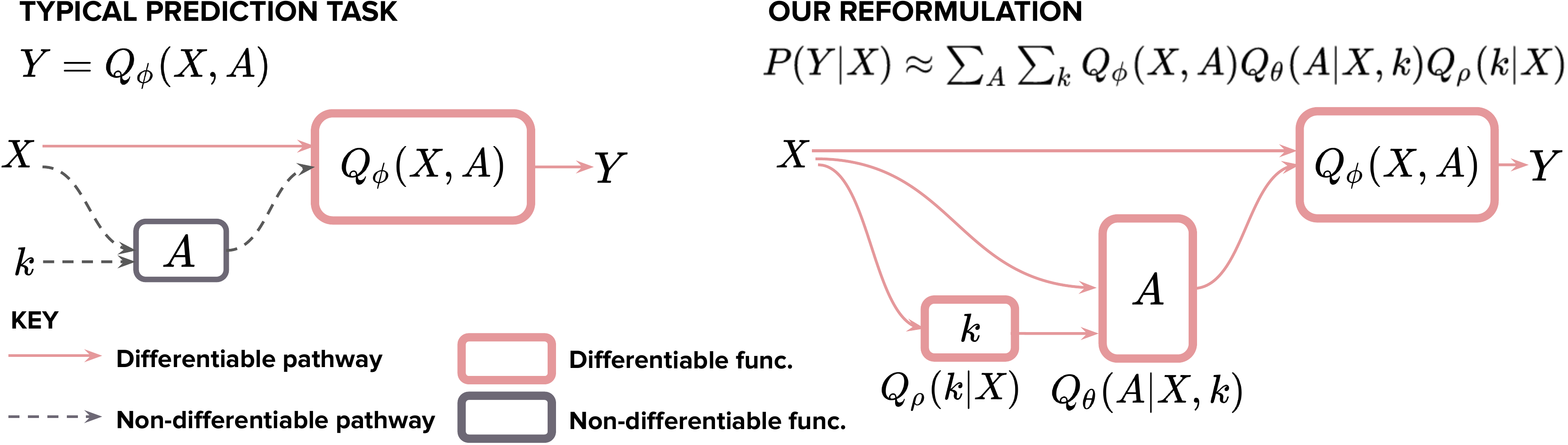}}
	\caption{(Left) A typical prediction task using graphs $Y = Q_\phi(X, A)$ where $A$ and $k$ are predetermined. (Right) Our reformulation $P(Y|X) \approx \sum_A \sum_k Q_\phi(X, A) Q_\theta(A|X,k) Q_\rho(k|X)$ which learns a distribution over $A$ and $k$ alongside the downstream task.\label{fig:task_overview}}

\end{figure}

\textbf{Problem definition} We  reformulate the baseline prediction task based on a fixed graph with an adaptive variant where the graph is learned.
Typically, such baseline tasks make learned predictions $Y$ given a set of input features $X$ and a graph structure $A$ of node degree $k$:
\begin{equation} \label{eq:baseline_task}
	Y = Q_\phi(X, A(k)),
\end{equation}
where $Q_\phi$ is an end-to-end neural network parameterized by learnable weights $\phi$. These formulations require a predetermined graph-structure $A(k)$, typically based on choice of  node degree $k$, and take $A(k)$ as additional input to the model. In contrast, we \emph{learn} both $A$ and $k$ in an end-to-end manner, and use them to make predictions $Y$.
 As graphs are inherently binary, with edges either present or absent, they are not directly optimizable using gradient descent. Instead, we  consider a distribution of graphs, $\cal G$, which then induce a distribution of labels, $\cal Y$, in the downstream task. This distribution takes the factorized form:
\begin{equation}
   P(Y|X) =
   \sum_{A\in \cal G} \sum_{k\in \mathbb{N}^{|V|}}
	Q_\phi(X, A) P(A|X,k)P(k|X),
\end{equation}
where $P(k|X)$ is the distribution of node degree $k$ given $X$ (i.e., the choice of $k$ in $k-$NN), $P(A|X,k)$ the distribution of graph structures $A$ conditioned on the learned $k$ and input $X$, and $P(Y|X)$ is the downstream distribution of labels
conditioned on data $X$. For clarity, the adjacency $A$ represents a subgraph of a complete graph over $X$, and $k$ is a multidimensional variable controlling the number of top-$k$ neighbors for each node individually. To avoid learning individual probabilities for each possible graph $A$ in an exponential state space, we further assume that $P(A|X,k)$ has a factorized distribution where each neighborhood is sampled independently, i.e. $P(A|X,k)=\prod_{i\in V} P(a_i|X,k)$.


We model the distributions over adjacencies $A$ and $k$ with tractable functions:
\begin{equation} \label{eq:tractable_funcs}
	P(Y|X) \approx \sum_A \sum_k Q_\phi(X, A) Q_\theta(A|X,k) Q_\rho(k|X),
\end{equation}
where $Q_\theta$ and $Q_\rho$ are functions parameterized by $\theta$ and $\rho$ to approximate  $P(A|X,k)$ and $P(k|X)$,  respectively. In Fig. \ref{fig:task_overview}, we illustrate the functions of our method compared to the typical prediction task in Eq. \ref{eq:baseline_task}.

Using this formulation, we train the entire system end-to-end to minimize the expected loss when sampling $Y$. This can be efficiently performed using stochastic gradient descent. In the forward pass, we first sample a subgraph/set of nodes $X$ from the space of datapoints, and conditioning on $X$ we sample $A$ and compute the associated label $Y$. When computing the gradient step, we update $ Q_\phi(X, A)$ as normal and update the distributions using two standard reparametrization tricks: one for discrete variables \cite{jang2016categorical} such that $Q_\theta(A|X,k)$ can generate differentiable graph samples $A'$, and another for continuous variables \cite{kingma2013auto} of $k'$ drawn from $Q_\rho(k|X)$:
\begin{equation}\label{eq:sampling_funcs}
\begin{split}
	P(Y|X) \approx \sum_{A'} \sum_{k'} Q_\phi(X, A'), \\
 	\textrm{where}~ A' \sim  Q_\theta(A|X,k') ~ \textrm{and} ~ k' \sim  Q_\rho(k|X).    
\end{split}
\end{equation}
As both the graph structure $A'$ and variable $k'$ samplers are differentiable, our DGG module can be readily integrated into pipelines involving graph convolutions and jointly trained end-to-end.

\subsection{Differentiable Graph Generation}
Our differentiable graph-generator (DGG) takes a set of nodes $V = \{v_1, ..., v_N\}$ with $d$-dimensional features $\mathbf{X} \in \mathbb{R} ^{N \times d}$  and generates a (potentially) asymmetric adjacency matrix $\mathbf{A} \in \mathbb{R}^{N \times N}$. This adjacency matrix can be used directly in any downstream graph convolution operation  (see Module Instantiation below).
As illustrated by Fig. \ref{fig:dgg}, the DGG module consists of four components:

\begin{figure}[t]
	\centerline{\includegraphics[page=4,trim=100 350 50 270,clip,width=1\linewidth]{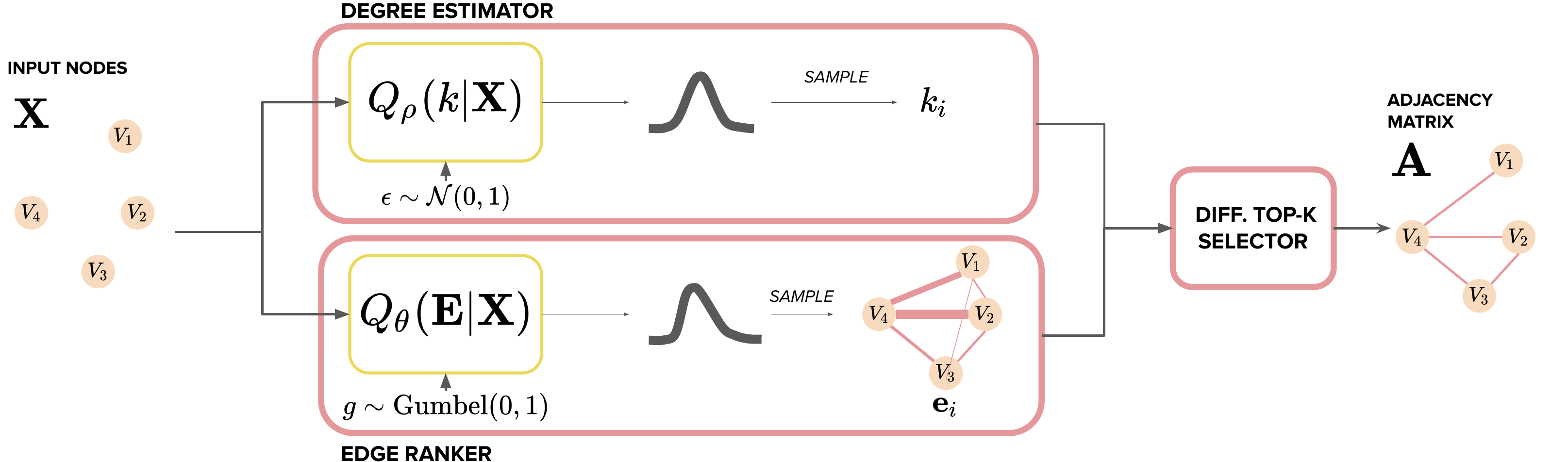}}
	\caption{Our differentiable graph-generator (DGG) takes input nodes $\mathbf{X}$ and generates an adjacency matrix $\mathbf{A}$. It consists of: (1) \textbf{Degree-estimator}: generates samples of $k_i$ for each node, (2) \textbf{Edge-ranker}: generates edge samples $\mathbf{e}_i$ for each node and (3) \textbf{Top-k selector}: takes $k_i$ and edge samples $\mathbf{e}_i$ and selects top-k elements in a differentiable manner to output a final adjacency $\mathbf{A}$. \label{fig:dgg}}
\end{figure}

\begin{enumerate}[nolistsep]
    
	\item \textbf{Node encoding:} this component projects the input node features $\mathbf{X} \in \mathbb{R} ^{N \times d}$ to a latent representation $\mathbf{\hat{X}} \in \mathbb{R} ^{N \times d'}$, which forms the primary representation space of the model.
    
	\item \textbf{Edge ranking}: this takes the latent node features $\mathbf{\hat{X}} \in \mathbb{R} ^{N \times d'}$ and generates a matrix representing a stochastic ordering of edges
	$\mathbf{E} \in \mathbb{R}^{N \times N}$ drawn from a learned distribution over the edge-probabilities ($A' \sim  Q_\theta(A|X,k')$ from Eq. \ref{eq:sampling_funcs}).
    
	\item \textbf{Degree estimation}: this component estimates the number of neighbors each individual node is connected to. It takes as input the latent node features $\mathbf{\hat{X}} \in \mathbb{R} ^{N \times d'}$ and generates random samples $k \in \mathbb{R}^N$ drawn from a learned distribution over the node degree ($k' \sim  Q_\rho(k|X)$ from Eq. \ref{eq:sampling_funcs}).
	\item \textbf{Differentiable top-\textit{k} edge selector}: takes $k$ and the edge-samples $e$ and performs a soft thresholding that probabilistically selects the most important elements, based on the output of the Edge-ranking 
	to output an adjacency matrix $\mathbf{A} \in \mathbb{R}^{N \times N}$.

\end{enumerate}
\color{black}
We now explain these steps in more detail:
 \\
\textbf{Node encoding} We construct a single latent space from the input node features, and use it for edge ranking and degree estimation. We first map input node features $\mathbf{X} \in \mathbb{R}^{N \times d}$ into latent features $\mathbf{\hat{X}} \in \mathbb{R}^{N \times d'}$ using a multi-layer perceptron (MLP) $N_\phi$ with weights $\phi$: $\hat{\mathbf{X}} = N_\phi(\mathbf{X})$. These latent features form the input for the rest of the DGG. Furthermore, they are output by the DGG and passed to the GCN downstream to prevent vanishing gradients.

\noindent \textbf{Edge ranking} The edge ranking returns an implicit distribution of edge orderings, from which we sample the neighborhood for each node. For each node $v_i$, it draws a set of scores $\mathbf{e}_i = \{e_{ij}\}_j^N$ quantifying its relevance to all nodes $v_j \in V$, including itself. To generate differentiable edge samples $\mathbf{e}_i$, we use the Gumbel-Softmax \cite{jang2016categorical}.

Before locally scoring each edge embedding $e_{ij} \in \mathbf{e}_i$ for node $v_i$, we implement a global stage which constructs edge embeddings with both local and global dependencies:
\begin{enumerate}[nolistsep]
	\item Using latent node features $\hat{\mathbf{x}}_i \in \hat{\mathbf{X}}$, determine local edge embeddings $\hat{\mathbf{c}}_{ij} \in \mathbb{R}^{d'}$ by passing each pair of node features through an MLP $l_\phi$: $\hat{\mathbf{c}}_{ij} = l_\phi(\hat{\mathbf{x}}_i, \hat{\mathbf{x}}_j)$. These embeddings now form a complete graph $\mathcal{G}$ over the nodes, with each edge attributed $\hat{\mathbf{c}}_{ij}$.
    
	\item As each edge embedding $\hat{\mathbf{c}}_{ij} \in \mathbf{C}$ is calculated independently of the others, we refine it to account for its dependencies to adjacent edges. We do this through edge-to-edge message passing. However, we avoid computing dependencies between all edges of the complete graph for two reasons: first, some edges may not have any common nodes, so passing messages between them could propagate irrelevant information, and secondly, it could be prohibitely expensive. To restrict message-passing between adjacent edges only, we first compute the adjoint graph $\mathcal{H}$ of the complete graph $\mathcal{G}$. In the adjoint $\mathcal{H}$, each edge is associated with a node, and two nodes are connected if and only if their corresponding edges in $\mathcal{G}$ have a node in common. The adjoint's adjacency $A^{\mathcal{H}}$ can be calculated using its incidence matrix $L$, $A^{\mathcal{H}} = L^TL - 2I$. In the adjoint, each node embedding $\hat{\mathbf{c}}_{i}$ is then updated using an average of its neighboring nodes $\hat{\mathbf{c}}_{j}$ and passed through an MLP $h_\phi$:
	\begin{equation}
    	\hat{\mathbf{c}}_{i}' = \sum_{j \in \mathcal{N}(i)} h_\phi(\hat{\mathbf{c}}_{i} \mathbin\Vert \mathbf{c}_{i} - \mathbf{c}_{j})
	\end{equation}
	\vspace{-0.5cm}
\end{enumerate}

Having computed edge embeddings $\mathbf{C} \in \mathbb{R}^{N \times N \times d'}$ with global dependencies, we rank these edges for each node. Without loss of generality, we focus on a single node $v_i \in V$, with latent features $\mathbf{\hat{x}}_i \in \mathbb{R}^d$. We implement the approximation function $Q_\theta(A|X,k)$ of the Edge-ranker as follows: \color{black}
\begin{enumerate}[nolistsep]

	\item \color{black}Using edge embeddings $\hat{\mathbf{c}}_{ij} \in \mathbb{R}^{d'}$, calculate edge probabilities $\mathbf{p}_i \in \mathbb{R}^N$ for node $v_i$ using an MLP $m_\theta$:
    	\begin{equation} \label{eq:dist}
        	\mathbf{p}_i = \{ m_\theta( \hat{\mathbf{c}}_{ij}) | \forall j \in N \}.
    	\end{equation}
    	Each element $p_{ij} \in \mathbf{p}_i$ represents a similarity measure between the latent features of node $v_i$ and $v_j$. In practice, any distance measure can be used here.\color{black}
   	 
	\item Using Gumbel-Softmax over the edge probabilities $\mathbf{p}_i \in \mathbb{R}^N$, we generate differentiable samples $\mathbf{e}_i \in \mathbb{R}^N$ with Gumbel noise $g$:
    	\begin{equation} \label{eq:softmax}
    	\begin{split}
                    	\mathbf{e}_i = \left\{ \frac{\exp((\log(p_{ij}) + g_i) + \tau)}{\sum_j \exp((\log(p_{ij}) + g_i) + \tau)} \bigg| \forall j \in N \right\}, \\
                    	g_i \sim \mathrm{Gumbel}(0,1)
    	\end{split}
    	\end{equation}
    	where $\tau$ is a temperature hyperparameter controlling the interpolation between a discrete one-hot categorical distribution and a continuous categorical density. When $\tau \rightarrow 0$, the edge energies $e_{ij} \in \mathbf{e}_i$ approach a degenerate distribution. The temperature $\tau$ is important for inducing sparsity, but given the exponential function, this results in a single element in $\mathbf{e}_i$ given much more weighting than the rest, i.e., it approaches a one-hot argmax over $\mathbf{e}_i$. As we want a variable number of edges to be given higher importance and others to be close to zero, we select a higher temperature and use the top-$k$ selection procedure (detailed below) to induce sparsity. This additionally avoids the high-variance gradients induced by lower temperatures.
\end{enumerate}

\noindent \textbf{Degree estimation} A key limitation of existing graph generation methods \cite{kazi2020differentiable,kipf2018neural,zheng2020robust} is their use of a fixed node degree $k$ across the entire graph. 
This can be suboptimal as mentioned previously.
In our approach, rather than fixing $k$ for the entire graph, we sample it per node from a learned distribution. 
Focusing on a single node as before, the approximation function $Q_\rho(k|X)$ of the Degree-estimator works as follows:
\begin{enumerate}[nolistsep]

	\item We approximate the distribution of latent node features $\hat{\mathbf{x}_i} \in \mathbb{R}^d$ following a  VAE-like formulation  \cite{kingma2013auto}. We encode its mean $\boldsymbol \mu_i \in \mathbb{R}^d$ and variance $\boldsymbol \sigma_i \in \mathbb{R}^d$ using two MLPs $M_\rho$ and $S_\rho$, and then reparametrize with noise $\epsilon$ to obtain latent variable $\mathbf{z}_i \in \mathbb{R}^d$:
	\begin{equation} \label{eq:degree_est_vae}
	\begin{split}
    	\boldsymbol{\mu}_i, \boldsymbol{\sigma}_i & = M_\rho(\hat{\mathbf{x}_i}), S_\rho(\hat{\mathbf{x}_i}), \\
    	\mathbf{z}_i & = \boldsymbol{\mu}_i + \boldsymbol{\epsilon}_i \boldsymbol{\sigma}_i, \epsilon_i \sim \mathcal{N}(0, 1).
	\end{split}
	\end{equation}
	\item Finally, we concatenate each latent variable $\mathbf{z}_i \in \mathbb{R}^d$ with the L1-norm of the edge samples $\mathbf{h}_i = ||\mathbf{e}_{i}||_1$ and decode it into a scalar $k_i \in  \mathbb{R}$ with another MLP $D_\rho$, representing a continuous relaxation of the neighborhood size for node $v_i$:
	\begin{equation}
    	k_i = D_\rho(\mathbf{z}_i) + \mathbf{h}_i.
	\end{equation}
   Since $\mathbf{h}_i$ is a summation of a node's edge probabilities, it can be understood as representing an initial estimate of the node degree which is then improved by combining with a second node representation $\mathbf{z}_i$ based entirely on the node's features. Using the edge samples to estimate the node degree links these representation spaces back to the primary latent space of node features $\hat{\mathbf{X}}$.
\end{enumerate}

\begin{figure*}[htb]
	\centerline{\includegraphics[page=3,trim=0 420 190 40,clip,width=0.6\linewidth]{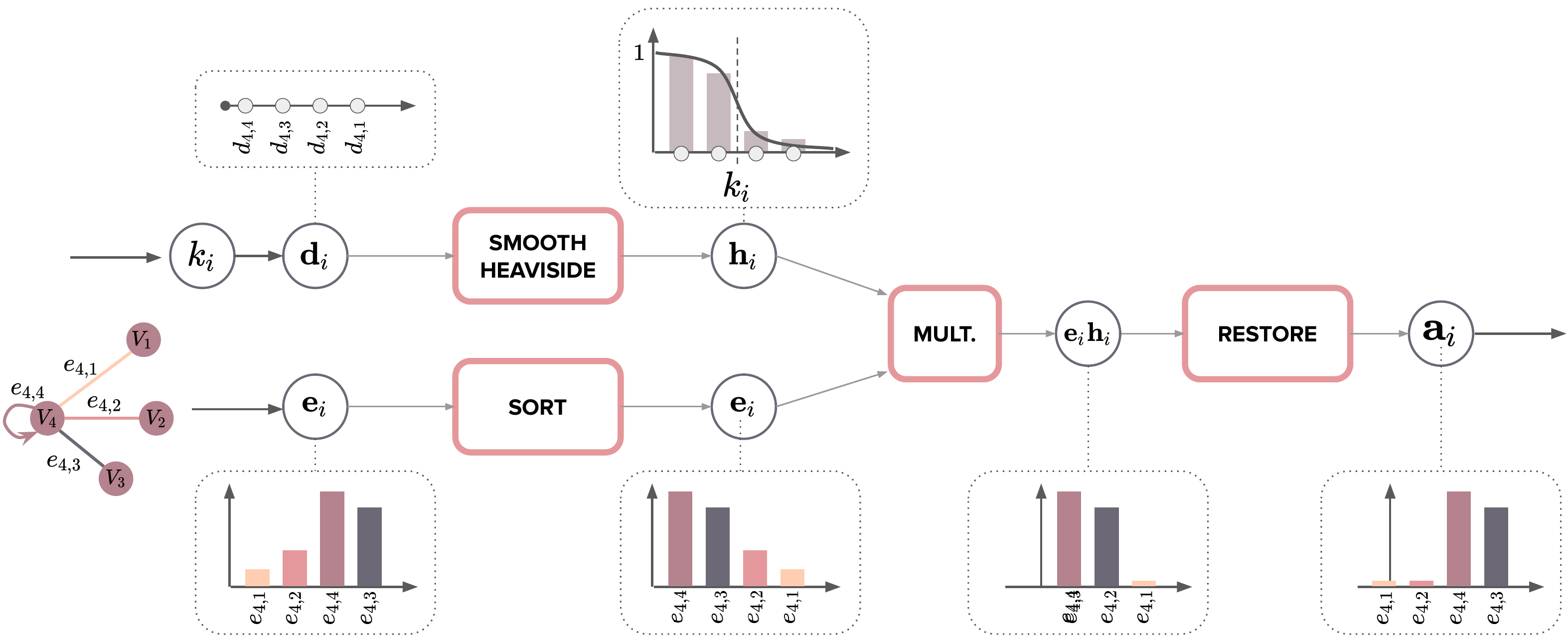}}
	\caption{The differentiable Top-$k$ Edge Selector. This component uses the node degree $k_i$ output by the Degree Estimator to control the inflection point on a smooth-Heaviside function and uses it to select the top edges from $\mathbf{e}_i$ output by the Edge Ranker. This produces an adjacency vector $\mathbf{a}_i$ for each node, and  stacking $\mathbf{a}_i$ across all nodes produces the final adjacency matrix $\mathbf{A}$. \label{fig:top_k_selector}}

\end{figure*}

\noindent \textbf{Top-\textit{k} Edge-Selector} Having sampled edge weights, and node degrees $k$, this function selects the top-$k$ edges for each node.
The top-$k$ operation, i.e. finding the indices corresponding to the $k$ largest elements in a set of values, is a piecewise constant function and cannot be directly used in gradient-based optimization.
Previous work \cite{xie2020differentiable} framed the top-$k$ operation as an optimal transport problem, providing a smoothed top-$k$ approximator. However, as their function is only defined  for discrete values of $k$ it cannot be optimized with gradient descent. As an alternative that is differentiable with respect to $k$, we relax the discrete constraint on $k$, and instead use it to control the $x$-axis value of the inflection point on a smoothed-Heaviside function (Fig. \ref{fig:top_k_selector}). For a node $v_i \in V$, of smoothed degree $k_i \in \mathbb{R}$ and edges $\mathbf{e}_i \in \mathbb{R}^N$, our Top-$k$ Edge Selector outputs an adjacency vector $\mathbf{a}_i \in \mathbb{R}^N$ where the $k$ largest elements from $\mathbf{e}_i$ are close to $1$, and the rest close to $0$. Focusing on a single node $v_i$ as before, the implementation is as follows:
\begin{enumerate}[nolistsep]

	\item Draw 1D input points $\mathbf{d}_{i} = \{1, ..., N\}$ where $N$ is the number of nodes in $V$.
    
	\item Pass $\mathbf{d}_{i}$ through a hyperbolic tangent (tanh) which serves as a smooth approximation of the Heaviside function:
	\begin{equation} \label{eq:smooth_heaviside}
    	\mathbf{h}_{i} = 1 - 0.5 * \left\{1 + \tanh (\lambda^{-1}d_i - \lambda^{-1}k_i) \right\},
	\end{equation}
	here $\lambda>0$ is a temperature parameter controlling the gradient of the function's inflection point. As $\lambda \rightarrow 0$,
	the smooth function approaches the Heaviside step function.
	The first-$k$ values in $\mathbf{h}_{i} = \{h_{ij}\}_j^N$ will now be closer to 1,
	while the rest closer to 0.
    
	\item Finally, for each node $i$ we sort its edge-energies $\mathbf{e}_i = \{e_{ij}\}_j^N$ in descending order, multiply by $\mathbf{h}_{i} = \{h_{ij}\}_j^N$ and then restore the original order to obtain the final adjacency vector $\mathbf{a}_i = \{a_{ij}\}_j^N$. Stacking $\mathbf{a}_i$ over all nodes $v_i \in V$ creates the final adjacency matrix $\mathbf{A} \in \mathbb{R}^{N \times N}$.

\end{enumerate}

\color{black}
\noindent \textbf{Symmetric adjacency matrix} If the adjacency matrix $A$ must be symmetric, this can be enforced by replacing it with $A_{sym}$ where: $\mathbf{A}_{sym} = (\mathbf{A} + \mathbf{A}^T)/2$.\color{black}

\noindent \textbf{Straight through Top-$k$ Edge Selector} To make our final adjacency matrix $\mathbf{A} \in \mathbb{R}^{N \times N}$ discrete, we follow the trick used in the Straight-Through Gumbel Softmax \cite{jang2016categorical}: we output the discretized version of $\mathbf{A}$ in the forward pass and the continuous version in the backwards pass. For the discretized version in the forward pass, we replace the smooth-Heaviside function in Eq. \ref{eq:smooth_heaviside} with a step function.

\textbf{Module Instantiation:}
The DGG module can be easily combined with any graph convolution operation. A typical graph convolution \cite{DBLP:conf/iclr/KipfW17} is defined as follows: $\mathbf{X}' = \hat{\mathbf{D}}^{-1/2} \hat{\mathbf{A}}\hat{\mathbf{D}}^{-1/2} \mathbf{X} \mathbf{\Theta}$.
Here, $\hat{\mathbf{A}} = \mathbf{A} + \mathbf{I}$ denotes the adjacency matrix with inserted self-loops, $\hat{\mathbf{D}}$ its diagonal degree matrix and $\mathbf{\Theta}$ its weights. To use this graph convolution with the DGG, we simply use our module to generate the adjacency matrix $\hat{\mathbf{A}}$.

\section{Experiments}
\color{black} We evaluate our DGG on node classification, point cloud classification and trajectory prediction. We chose these tasks as they demonstrate the wide applicability of our module: (1) graphs for node classification require models that can generate edge structures from  noisy input graphs, (2) point cloud classification tasks have no input graph structures and (3) trajectory prediction additionally requires models which can handle a variable number of nodes per batch. We compare against state-of-the-art structure learning methods in each domain. As far as we know, our structure-learning approach is the only one that can be easily applied without modification to any GCN pipeline in such a range of tasks.


\subsection{Node classification}
Beginning with node classification, we conduct ablations examining the behavior of different parts of the DGG, followed by comparisons to other state-of-the-art structure learning approaches. In the supplementary we include experiments investigating the effect of the DGG on downstream models under the addition of noisy edges to input graphs. We perform these experiments under both transductive and inductive scenarios, as well as semi-supervised and fully-supervised settings.

\textbf{Datasets} In the transductive setting, we evaluate on three citation benchmark datasets Cora, Citeseer and Pubmed \cite{pubmed_nlm} introduced by \cite{yang2016revisiting}. In an inductive setting, we evaluate on Reddit \cite{graphsaint-iclr20}  and PPI \cite{hamilton2017inductive}. Further dataset details can be found in the supplementary.
\textbf{Baselines and Implementation} \color{black} As our DGG is a GCN-agnostic module that can be integrated alongside any graph convolution operation, we compare its performance to both other GCN-agnostic approaches and bespoke structure-learning architectures. To compare against other GCN-agnostic methods, we int
egrate our DGG into four representative GCN backbones: GCN \cite{DBLP:conf/iclr/KipfW17}, GraphSage \cite{hamilton2017inductive}, GAT \cite{velivckovic2018graph} and GCNII \cite{chen2020simple}. On these backbones, we compare against other GCN-agnostic structure learning methods: DropEdge \cite{rong2020dropedge}, NeuralSparse \cite{zheng2020robust}, PTDNet \cite{luo2021learning}.  Then we compare against bespoke architectures IDGL \cite{chen2020iterative}, LDS \cite{franceschi2019learning}, SLAPS \cite{fatemi2021slaps}, NodeFormer \cite{wu2022nodeformer} and VGCN \cite{elinas2020variational}. To make our comparison fair against these bespoke architectures which learn the structure specifically for node classification, we integrate our DGG into a GCN backbone that is comparable to the bespoke architecture in design. Please see the supplementary for implementation details.\color{black}

\textbf{Training details} A node classification model partitions the latent space of node embeddings into separate classes. However, when message-passing, there is one phenomenon of the input graph that can limit classification accuracy: two nodes with different classes but similar features and an edge connecting them. Classifying these nodes is challenging as their feature similarity can be compounded by passing messages between them.
The goal of the DGG is to move such nodes apart in the latent space such that there is no edge and communication between them. However, traversing the loss landscape from the initial random initialization of the network to one where the model is able to discriminate between these nodes can take several iterations using only the downstream classification loss. To speed up training, we add an intermediate loss to further partition the latent space. We do this by supervising the adjacency matrix generated by the DGG to remove all edges between classes and only maintain those within a class. We then anneal this loss over the training cycle, eventually leaving only the downstream classification loss. We provide more details in the supplementary.

\subsubsection{Ablations}
In Table \ref{tab:ablations}, we explore the effect of disabling different components of our DGG module when integrated into a GCN \cite{DBLP:conf/iclr/KipfW17} for node classification: 1. \textit{DGG without Degree Estimation and Differentiable Top-$k$ Edge Selection} --- we remove the Degree Estimator and instead fix $k$ to select the top-$k$ stochastically ordered edges. 2. \textit{DGG with deterministic Edge Ranking} --- we remove the noise in Eq. \ref{eq:softmax} of the Edge Ranker. 3. \textit{DGG with deterministic Degree Estimation} --- we remove the noise in Eq. \ref{eq:degree_est_vae} of the Degree Estimator.
We perform these
on Cora \cite{yang2016revisiting} and omit the annealed intermediate loss during training.

\begin{table}[htb]
\centering
\caption{Ablation study. DGG integrated into a GCN for node classification on Cora \cite{yang2016revisiting}.} 
\label{tab:ablations}
\resizebox{\columnwidth}{!}{%
\begin{tabular}{cc}
\hline
Model & Accuracy \\ \hline
Fixed node degree, k = \{1, 5, 10, 100\} & \{49.7, 78.9, 55.0, 37.0\} \\
With deterministic Edge Ranking and Degree Estimation & 82.4 \\
With deterministic Edge Ranking & 82.7 \\
With deterministic Degree Estimation & 82.8 \\
DGG & \bf 83.2 \\ \hline
\end{tabular}
}
\vspace{-0.35cm}
\end{table}

Table \ref{tab:ablations} shows the benefit of learning a distribution over the node degree. When learning it deterministically, the accuracy decreases by 0.5\%. This becomes significantly worse when the node degree is fixed for the entire graph rather than learned per node. Note also, the sensitivity with respect to choice of $k$. A fixed node degree of $k = 10$ or $k=1$ reduces accuracy by almost 30\% vs a graph of 5. This is due to the graph convolution operation: as it has no adaptive weighting mechanism for a node's neighborhood, each of the  neighbors is given the same weight. Naturally, this leads to information sharing between unrelated nodes, reducing the quality of node representation after message-passing. In contrast, by learning a distribution over the node degree we are able to select only the most relevant neighbors, even though these are then weighted equally in the graph convolution. Finally, the inclusion of noise in any of the DGG components does increase accuracy, but only by approximately 0.5\% --- demonstrating both its benefit and the robustness of the DGG without it.

\subsubsection{Results}
\textbf{Comparison to GCN-agnostic modules} In Table \ref{tab:node_classification_vs_sota} we compare against GCN-agnostic structure learning methods. For fair comparison
, we present two versions of our method: DGG-wl trained with the downstream loss only and DGG* trained with both the downstream and intermediate loss.

\begin{table}[htb]
\centering
\caption{Semi-supervised node classification compared to other \textit{architecture agnostic} SOTA structure learning methods. We compare against prior methods reported in \cite{luo2021learning, zheng2020robust, chen2020simple}, using the official results where available. Those with `-' have no official results or we ran into out-of-memory errors.}
\resizebox{\columnwidth}{!}
{%
\label{tab:node_classification_vs_sota}
\begin{tabular}{c|l|ccccc} \toprule
{\textbf{Backbone}}               	& \textbf{Method}   	& \textbf{Cora}      	& \textbf{Citeseer}  	& \textbf{Pubmed} 	& \textbf{Reddit} 	& \textbf{PPI}   \\ \bottomrule
\multirow{5}{*}{\begin{sideways}GCN\end{sideways}}                        	& Original 	& 81.1      	& 70.3      	& 79.0   	& 92.2     & 53.2   \\
                                            	& DropEdge 	& 80.9      	& 72.2      	& 78.5  	& 96.1    & 54.8	\\
                                            	& NeuralSparse & 82.1      	& 71.5      	& 78.8  	& 96.6    & 65.1	\\
                                            	& PTDNet-wl  	& 82.4 & 71.7 & 79.1 & - & 75.2\\
                                            	& DGG-wl     	& \textbf{83.2}      	& \textbf{72.6}      	& \textbf{80.2} & \textbf{96.8}	& \textbf{77.1}  	\\ \cline{2-7}
                                            	& PTDNet-wl + low rank   	& 82.8 & 72.7 & 79.8 & - & 80.3\\
                                            	& DGG*     	& \underline{\textbf{84.1}}      	& \underline{\textbf{74.9}}      	& \underline{\textbf{84.0}} & \underline{\textbf{97.3}}	& \underline{\textbf{81.6}}  	\\ \bottomrule
\multirow{5}{*}{\begin{sideways}GraphSage\end{sideways}} & Original 	& 79.2      	& 67.6      	& 76.7 	& 93.8 & 61.8 	\\
{}                       	& NeuralSparse & 79.3      	& 67.4      	& 75.1   & \textbf{96.7} & 62.6   	\\
{}                       	& PTDNet-wl   	& \textbf{79.4}      	& 67.8      	& 77.0   & -	& 64.5   	\\
{}                       	& DGG-wl     	& \textbf{79.4}  & \textbf{68.2}  & \textbf{77.6}  & 96.6	& \textbf{65.3}\\ \cline{2-7}
{}                       	& PTDNet-wl + low rank   	& 80.3      	& 67.9      	& 77.1   & -	& 64.8   	\\
{}                       	& DGG*     	& \underline{\textbf{80.5}}  & \underline{\textbf{70.8}}  & \underline{\textbf{80.2}}  & \underline{\textbf{96.9}}	& \underline{\textbf{67.3}}\\ \bottomrule
\multirow{5}{*}{\begin{sideways}GAT\end{sideways}}                        	& Original 	& 83.0      	& 72.1      	& 79.0 	& -   &   97.3 	\\
                                            	& DropEdge 	& 83.2      	& 70.9      	& 77.9   & -   &   85.1   	\\
                                            	& NeuralSparse & 83.4      	& 72.4      	& 78.0   & -   &   92.1   	\\
                                            	& PTDNet-wl   	& 83.7      	& 72.3      	& 79.2   & -   &   \textbf{97.8}   	\\
                                            	& DGG-wl     	& \textbf{84.6}  & \textbf{73.2}  & \textbf{79.7}  & -   & 97.4\\ \cline{2-7}
                                            	& PTDNet-wl + low rank   	& 84.4      	& 73.7      	& 79.3   & -   &   \underline{\textbf{98.0}}   	\\
                                            	& DGG*    	& \underline{\textbf{85.3}}  & \underline{\textbf{76.4}}  & \underline{\textbf{82.0}}  & -   & 97.6\\ \bottomrule
\multirow{5}{*}{\begin{sideways} GCNII \end{sideways}}                        	& Original 	& 85.3      	& 73.2      	& 80.2 	& -   &   99.5 	\\
                                            	& DropEdge 	& 84.9      	& 73.4      	& 79.4   & -   &   99.0   	\\
                                            	& DGG-wl     	& \textbf{86.9}  & \textbf{74.5}  & \textbf{81.5}  & -   & \textbf{99.6} \\
                                            	& DGG*    	& \underline{\textbf{87.8}}  & \underline{\textbf{75.7}}  & \underline{\textbf{81.9}}  & -   & \underline{\textbf{99.7}}\\ \bottomrule
\end{tabular}
}

\end{table}

DGG improves performance
across all baselines and datasets. Against other approaches, DGG-wl generally outperforms the state-of-the-art NeuralSparse and PTDNet-wl (both trained with only the downstream loss). This can be attributed to our method for modelling sparsity, which explicitly lets each node to select the size of its neighborhood based on the downstream training signal. This training signal helps partition the node representation space, while the estimated node-degree additionally prevents communication between distant nodes. Although PTDNet-wl does this implicitly through its attention mechanism, discovering this sparse subgraph of the input graph is challenging given its complexity. NeuralSparse on the other hand selects  $k$ for its entire generated subgraph, which is both suboptimal and  requires additional hyperparameter tuning.

\begin{table}[htb]
\centering
\caption{Adjacency matrix constraints: our intermediate annealed loss vs. PTDNet's low rank regularizer \cite{luo2021learning} for semi-supervised node classification with a GCN backbone.}
\label{tab:loss_comparisons}
\resizebox{\columnwidth}{!}{%
\begin{tabular}{l|lllll}
\hline
\textbf{Method} & \textbf{Cora} & \textbf{Citeseer} & \textbf{Pubmed}  & \textbf{PPI} \\ \hline
DGG-wl & 86.8 & 74.4 & 81.2 & 99.5 \\
DGG-wl + low rank & 87.1 & 75.3 & 81.4 & 99.5 \\
DGG-wl + int. loss (aka DGG*) & \textbf{87.7} & \textbf{75.8} & \textbf{81.9} & \underline{\textbf{99.7}} \\ \hline
DGG-wl + int. loss + low rank & \underline{\textbf{87.8}} & \underline{\textbf{76.2}} & \underline{\textbf{82.1}} & \underline{\textbf{99.7}} \\ \hline
\end{tabular}
}
\end{table}

Comparing methods which enforce additional constraints on the adjacency matrix, DGG* demonstrates larger accuracy gains than PTDNet*. PTDNet* regularizes its adjacency matrix to be of low-rank, as previous work \cite{savas2011clustered} has shown that the rank of an adjacency matrix can reflect the number of clusters. This regularizer reasons about the graph's topology globally. While this may aid generalization, the accuracy difference may then be attributed to our intermediate loss providing stronger signals to discriminate between nodes with similar features but different classes (and therefore remove the edges between them). Furthermore, their regularizer uses the sum of the top-$k$ singular values during training, where $k$ again is a hyperparameter tuned to each dataset individually. Our method requires no additional parameters to be chosen.


Finally in Table \ref{tab:loss_comparisons} we compare the low-rank constraint of PTDNet with our intermediate annealed loss. Our intermediate loss (`DGG-wl + int. loss') outperforms the low-rank constraint (`DGG-wl + low rank').
However, using both constraints (`DGG-wl + int. loss + low rank') increases classification accuracy further, suggesting the edges removed by both methods are complementary.

\begin{table}[]
\centering
\color{black}
\caption{Node classification results against bespoke SOTA architectures which learn the graph structure.}
\label{tab:node_classification_sota}
\resizebox{\columnwidth}{!}{%
\begin{tabular}{l|l|cc}
\hline
      	& \textbf{Model}  	& {\textbf{Cora}} & {\textbf{Citeseer}} \\ \hline
\textbf{Setting 1} & IDGL \cite{chen2020iterative}  	& 84.5                 	& 74.1                     	\\
	original  	& DGG* + GAT  & \textbf{85.4}        	& \textbf{76.4}            	\\ \hline
\textbf{Setting 2} & LDS \cite{franceschi2019learning}   	& 71.5                 	& 73.3                     	\\ Input graph = none
      	& SLAPS \cite{fatemi2021slaps}  	& 74.2                 	& 73.1                     	\\ train split = \{train + 1/2 val\}
      	& DGG* + GCN  & \textbf{82.4}        	& \textbf{74.0}            	\\ \hline
\textbf{Setting 3} & NodeFormer \cite{wu2022nodeformer} & 88.7                 	& 76.2                     	\\
   split = 0.5/0.25/0.25   	& DGG* + GAT  & \textbf{90.1}        	& \textbf{77.8}            	\\ \hline
\textbf{Setting 4} & VGCN \cite{elinas2020variational}   	& 85.9                 	& 76.5                     	\\
   train split = \{train + 1/2 val\}   	& DGG* + GAT  & \textbf{87.6}        	& \textbf{77.1}            	\\ \hline
   \color{black}
\end{tabular}%
}
\vspace{-0.15cm}
\end{table}

\color{black}
\textbf{Comparison with bespoke architectures} In Table \ref{tab:node_classification_sota} we compare against bespoke architectures specifically designed for node classification. As each of these methods uses different experiment settings, we train our DGG-integrated architecture separately for each. See the supplementary for details on each setting and reasons for our choice of backbone. Our performance gains here can generally be attributed to factors: (1) our intermediate loss on the adjacency matrix and (2) our adjacency matrix factorizations where we learn the neighborhood for each node. Our intermediate loss particularly benefits from the experimental settings adopted by the other methods as they use larger training splits involving half the validation graph. Additionally, constructing the adjacency matrix by learning nodewise neighborhoods restricts the graph search space, making optimization easier. However, we note that some of these other methods are designed for node-classification on graphs which are orders of magnitude larger than Cora and Citeseer. In such cases, factorizing the adjacency per node, as we do, may be unfeasible.
\color{black}

\begin{table}[tbp]
\centering
\caption{ADE/FDE on the ETH \& UCY datasets using Social-STGCNN (first table), and Stanford Drone Dataset (SDD) using DAGNet (second table). For DGM \cite{kazi2020differentiable}, $k = 2$ for both datasets. }
\label{tab:results_pedestrian_traj_pred}
\resizebox{\columnwidth}{!}
{%
\begin{tabular}{ccccccccccc}
\toprule
\textbf{Dataset} & \multicolumn{2}{c}{\textbf{Original}}  & \multicolumn{2}{c}{\textbf{DGM \cite{kazi2020differentiable} Gain (\%)}} & \multicolumn{2}{c}{\textbf{DGG Gain (\%)}}  \\
                               	& ADE$\downarrow$      	& FDE $\downarrow$    	& ADE $\uparrow$    	& FDE$\uparrow$ & ADE $\uparrow$    	& FDE $\uparrow$    	\\ \toprule
ETH                  	& 0.64      	& 1.11          	& 2.4\%     	& 6.4\%             	& \textbf{7.8\%} & \textbf{21.4\%} \\
Hotel                	& 0.49      	& 0.85          	& 14.2\%    	& 18.9\%             	& \textbf{22.7\%} & \textbf{37.5\%} \\
Univ                 	& 0.44      	& 0.79           	& 6.2\%    	& 3.5\%               	& \textbf{11.8\%}     	& \textbf{14.9\%}     	\\
Zara1                	& 0.34      	& 0.53           	& 3.8\%   	& 13.7\%               	& \textbf{7.7\% }    	& \textbf{23.8\%}     	\\
Zara2                	& 0.30      	& 0.48          	& 5.0\%   	& 5.0\%              	& \textbf{12.8\%}   	& \textbf{17.3\%}     	\\
Mean              	& 0.44      	& 0.75           	& 6.3\%    	& 10.6\%               	& \textbf{12.6\%}    	& \textbf{23.0\%}     	\\ \bottomrule
 SDD & 0.53      	& 1.04         	& 1.9\%     	& 3.0\%     	& \textbf{10.9\%}     	& \textbf{9.5\%} \\ \bottomrule
\end{tabular}
}
\end{table}

\begin{table}[tbp]
\centering
\caption{ADE/FDE metrics on the SportVU Basketball dataset using DAGNet. For DGM \cite{kazi2020differentiable}, $k = 3$.}
\label{tab:results_dagnet}
\resizebox{\columnwidth}{!}
{%
\begin{tabular}{cccccccc}
\toprule
                                        	&                      	& \multicolumn{2}{c}{\textbf{Original}}                                  	& \multicolumn{2}{c}{\textbf{DGM \cite{kazi2020differentiable} Gain (\%)}}                                      	& \multicolumn{2}{c}{\textbf{DGG Gain (\%)}}                  	\\
\textbf{Split}                                  	& \textbf{Team}                 	& {ADE} & {FDE} & {ADE} & {FDE} & {ADE} & {FDE} \\ \toprule
{\multirow{2}{*}{10-40}} & {ATK} & 2.74                	& 4.29                                    	& -0.4\%                	& -0.2\%                              	& \textbf{6.7\%}                 	& \textbf{5.1\%}                 	\\
{}                   	& {DEF} & 2.09                	& 2.97                                    	& -0.5\%                	& -0.1\%                                  	& \textbf{9.7\%}                 	& \textbf{6.4\%}                 	\\ \hline
{\multirow{2}{*}{20-30}} & {ATK} & 2.03                	& 3.98                                    	& 0.1\%                 	& 0.1\%                             	& \textbf{7.2\%}                 	& \textbf{8.2\%}                 	\\
{}                   	& {DEF} & 1.53                	& 3.07                                    	& 0.2\%                 	& 0.3\%                                 	& \textbf{21.4\%}                	& \textbf{19.1\%}                	\\
\hline
{\multirow{2}{*}{40-10}} & {ATK} & 0.81                	& 1.71                                   	& 1.3\%                 	& 0.9\%                                   	& \textbf{15.5\%}                	& \textbf{17.0\%}                	\\
{}                   	& {DEF} & 0.72                	& 1.49                                    	& 0.8\%                 	& 0.8\%                                    	& \textbf{10.9\%}                 	& \textbf{16.2\%}                	\\
\hline
{Mean}            	& {---}	& 1.65                	& 2.92                                 	& 0.3\%                 	& 0.3\%                                    	& \textbf{11.9\%}                 	& \textbf{12.0\% }                	\\ \hline
\end{tabular}
}
\vspace{-0.6cm}
\end{table}


\subsection{Trajectory prediction}
\color{black} We evaluate on trajectory prediction tasks as these have neither an input or ground truth graph structure, thus the ideal structure has to be generated entirely from the data. \color{black} We consider four datasets covering a range of scenarios from basketball to crowded urban environments. On each, we integrate our DGG into a SOTA GCN trajectory prediction pipeline and compare results to another task-agnostic structure learning approach, DGM \cite{kazi2020differentiable}.

\textbf{Datasets}
We evaluate on four trajectory prediction benchmarks. 1. ETH \cite{pellegrini2009you} and UCY \cite{lerner2007crowds} --- 5 subsets of widely used real-world pedestrian trajectories. 2. STATS SportVU \cite{stats_perform_2019} --- multiple NBA seasons tracking trajectories of basketball players over a game. Stanford Drone Dataset (SDD) \cite{robicquet2016learning} --- top-down scenes across multiple areas at Stanford University. Further details on these datasets can be found in the supplementary. \textbf{Baselines and Implementation}
We integrate our DGG module into two state-of-the-art trajectory prediction pipelines: Social-STGCNN \cite{mohamed2020social} and DAGNet \cite{monti2021dag}. Our DGG is placed within both networks to generate the adjacency matrix on the fly and forms part of its forward and backward pass. Please see the supplementary for implementation details. \textbf{Evaluation metrics.}
Model performance is measured with Average Displacement Error (ADE) and Final Displacement Error (FDE). ADE measures the average Euclidean distance along the entire predicted trajectory, while the FDE is that of the last timestep only.

\textbf{Results} $\;$
In Table \ref{tab:results_pedestrian_traj_pred}, the integration of our DGG into Social-STGCNN reduces ADE/FDE compared to both the baseline and the integration of DGM.
In Table \ref{tab:results_pedestrian_traj_pred} and \ref{tab:results_dagnet} we demonstrate similar gains over DGM when integrated into DAGNet. First, this shows the benefit of inducing sparsity when message-passing over a distance weighted adjacency matrix like Social-STGCNN or even an attention-mechanism like DAGNet. The larger error reduction of our DGG compared to DGM may be attributed to DGM's use of a fixed node-degree $k$ across its learned graph. While this can prevent the propagation of irrelevant information across the graph in some cases, in others it might limit the context available to certain nodes.
We provide qualitative analysis in the supplementary.


\subsection{Point Cloud Classification}
\color{black}
We evaluate on another vision task of point cloud classification for models which use GCNs. This task differs from the previous two as predictions are made for the entire graph as opposed to node-wise. As with our trajectory prediction experiments, we integrate our DGG into SOTA classification architectures and compare against the other task-agnostic graph-learning module DGM \cite{kazi2020differentiable}.

\textbf{Datasets} We evaluate on ModelNet40 \cite{wu20153d}, consisting of CAD models for a variety of object categories.  \textbf{Baselines and Implementation} We integrate our DGG into a SOTA ResGCN \cite{li2021deepgcns} and DGCNN \cite{wang2019dynamic}. Both models use a $k$-NN sampling scheme to construct its graph. We simply replace this sampler with our DGG and keep the rest of the network and training protocol the same.

\textbf{Results} $\;$
Our results in Table \ref{tab:pt_classification_results} demonstrate the benefits of learning an adaptive neighborhood size across the latent graph. DGM \cite{kazi2020differentiable} learns a fully-connected latent graph and then imposes a fixed node degree of $k=20$ across it (i.e. selecting the top 20 neighbors for each node). This marginally improves upon the baselines ResGCN \cite{li2021deepgcns} and DGCNN\cite{wang2019dynamic}, which both also used fixed node-degrees $k$. In contrast, we learn a distribution over the node degree from which we sample each node's neighborhood size.  As shown in Table \ref{tab:pt_classification_results}, the node degree varies in our models with a standard deviation of around 5-7 across both baselines. Our accuracy gains over the baseline and DGM can be attributed to this variance in neighborhood sizes across the graph. These gains can be understood when viewing an input point cloud as a composition of object parts. Building semantic representations for different parts may naturally require varying amounts of contextual points. For instance, the wheels of a car might be identifiable with a smaller neighborhood than the car's body. This may suggest why an adaptive neighborhood size is helpful in this case.
\color{black}

\begin{table}[]
\centering
\color{black}
\caption{Point Cloud classification on ModelNet40 with our module and DGM \cite{kazi2020differentiable} integrated into two different point cloud labelling architectures.}
\label{tab:pt_classification_results}
\resizebox{\columnwidth}{!}{%
\begin{tabular}{l|l|cc|c}
\hline
\textbf{Baseline} & \textbf{Method} & \textbf{Mean degree} & \textbf{S.D. degree} & \textbf{Accuracy} \\ \hline
ResGCN \cite{li2021deepgcns}        	& Original    	& 9           	& 0          	& 93.3          	\\
              	& DGM \cite{kazi2020differentiable}         	& 20          	& 0          	& 93.5          	\\
              	& DGG        	& 14.8        	& 7.4        	& \textbf{94.4} 	\\ \hline
DGCNN \cite{wang2019dynamic}         	& Original    	& 40          	& 0          	& 92.9          	\\
              	& DGM \cite{kazi2020differentiable}           	& 20          	& 0          	& 93.3          	\\
              	& DGG        	& 19.3        	& 5.2        	& \textbf{93.8} 	\\ \hline
\end{tabular}%
}
\end{table}

\section{Conclusion}
We have presented a novel approach for learning graph topologies, and  shown how it obtains state-of-the-art performance across multiple baselines and datasets for trajectory prediction, point cloud classification and node classification. \color{black} The principal advantage of our approach is that it can be combined with any existing graph convolution layer, under the presence of noisy, incomplete or unavailable edge structures. \color{black} 
 \vspace{-0.5cm}
\section*{Acknowledgements}
This project was supported by the EPSRC project ROSSINI (EP/S016317/1) and studentship 2327211 (EP/T517616/1).

{\small
\bibliographystyle{ieee_fullname}
\bibliography{main.bib}
}

\end{document}